# Source-Aware Embedding Training on Heterogeneous Information Networks

Tsai Hor Chan
*Department of Statistics and Actuarial Science*
*The University of Hong Kong*
Hong Kong, China
hchanth@connect.hku.hk

Chi Ho Wong
*Department of Computer Science and Engineering*
*The Hong Kong University of Science and Technology*
Hong Kong, China
chwongcc@connect.ust.hk

Jiajun Shen
*TCL Corporate Research (Hong Kong)*
Hong Kong, China
shenjiajun90@gmail.com

Guosheng Yin
*Department of Statistics and Actuarial Science*
*The University of Hong Kong*
Hong Kong, China
gyin@hku.hk

*Abstract*—Heterogeneous information networks (HINs) have been extensively applied to real-world tasks, such as recommendation systems, social networks, and citation networks. While existing HIN representation learning methods can effectively learn the semantic and structural features in the network, little awareness was given to the distribution discrepancy of subgraphs within a single HIN. However, we find that ignoring such distribution discrepancy among subgraphs from multiple sources would hinder the effectiveness of graph embedding learning algorithms. This motivates us to propose SUMSHINE (Scalable Unsupervised Multi-Source Heterogeneous Information Network Embedding) --- a scalable unsupervised framework to align the embedding distributions among multiple sources of an HIN. Experimental results on real-world datasets in a variety of downstream tasks validate the performance of our method over the state-of-the-art heterogeneous information network embedding algorithms.

## I. INTRODUCTION

Heterogeneous information network (HIN), also known as heterogeneous graph, is an advanced graph data structure which contains enriched structural and semantic information. Learning the representations of HINs has recently drawn significant attention for its outstanding contribution to industrial applications and machine learning research.

HINs have a variety of real-world applications including recommendation systems [1], citation networks [2], natural language processing [3, 4], and social media [5, 6]. An HIN is a multi-relation and multi-entity graph summarizing the relations between entities, which represents a key abstraction for organizing information in diverse domains and modelling real-world problems in a graphical manner.

Heterogeneous information network embedding methods aim to encode each of the entities and relations in the HIN to a low-dimensional vector, which give feature representations to entities and relations in the HIN. Since the multi-relation and multi-entity characteristics introduce heterogeneity to HINs and feature different distributions among different types of entities and relations, state-of-the-art (SOTA) methods mostly focus on developing transformation techniques to bring feature distributions of different entity types and relation types to the same embedding space [3, 7, 8].

However, as of today, SOTA methods often operate on an HIN constructed by subgraphs from multiple sources, and most research has been based on the often implicit assumption that the effect of distribution discrepancies among different subgraphs on embedding learning is negligible. The major contribution of this work is to raise awareness to the graph learning community that this assumption does not hold in many cases. For instance, graph-based recommendation sys- tem often takes advantage of the information embedded in HINs, where an HIN often contains a user-content interaction graph with high-degree content entity nodes as well as a knowledge graph with low-degree content entity nodes. The difference in graph structures (i.e., average node degrees, graph sizes, sparsity of connections) leads to distribution discrepancies among subgraphs sources in the HIN. As we will show in this paper, simply ignoring such distribution discrepancies when training HIN embeddings would lead to sub-optimal embedding learning performance.

Although none of the existing heterogeneous graph embedding approaches attempt to solve the aforementioned problem, there are several attempts in heterogeneous graph neural networks (GNNs) that try to transfer a GNN model trained on one graph to another [9, 10]. They often apply domain transfer techniques to graph neural networks so that the knowledge learned from one graph can be better transferred to another. Note that these approaches differ from our approach in the following important aspects: 1) Unlike the supervised learning nature of GNN models, we are tackling the graph embedding learning task which aims to infer node representations from graph structures in an unsupervised manner. 2) These domain adaption approaches often focus on adapting the learned model of one graph to another, while we focus on how to learn one model from a graph merged from sources.

In this work, we study the distribution discrepancy is- sue in heterogeneous graph embedding learning. We sur-mise that simply merging sub-graphs from different sources when training graph embeddings may negatively impact the effectiveness, which unfortunately is *de facto* the only known approach to leverage data from multiple graphs. Motivated by this limitation, we develop a scalable unsupervised multi-source representation learning framework for learning heterogeneous information network embeddings, named SUMSHINE (**S**calable **U**nsupervised **M**ulti-**S**ource **H**eterogeneous **I**nformation **N**etwork **E**mbedding). It allows to train large-scale heterogeneous information network embeddings from different sources into a distribution-aligned latent embedding space, and we confirmed that the embedding learning performance can be significantly improved as our framework is designed to cope with the distribution discrepancy issue in learning heterogeneous information network embeddings.

Our contributions can be summarized as follows:

- We study the distribution misalignment problem in HIN embeddings and conclude that the HIN embeddings should be trained with distribution alignment performed on the subgraph sources of the HIN to achieve optimal downstream task performance. To the best of our knowledge, we are the first to introduce source-level distribution alignment to heterogeneous information network embedding.
- We propose source-aware negative sampling to balance the training samples by source, while preserving the scalability advantage of negative sampling. This design overcomes the scalability constraints of existing HIN embedding methods using GNNs.
- We validate our proposed method empirically on both link prediction and node classification downstream tasks, using a variety of real-world datasets. We also highlight a practical application of our method on recommendation systems with extensive experiments.

II. RELATED WORKS

*A. Heterogeneous Information Network Embedding*

Heterogeneous information network embedding has shown significant successes in learning the feature representations of an HIN. Existing HIN embedding methods aim to learn a low dimensional feature representation of an HIN. They apply different transformation techniques to bring the embeddings into the same latent embedding space [7, 8]. Most of the HIN embedding methods focus on leveraging the multi-relation characteristic in the HIN, which are known as similarity-based methods [3, 4, 11, 12, 13]. Similarity-based methods are widely adopted to learn the HIN representations by encoding the similarity between the source and destination entities in an edge. Within this class, there are translational methods, such as TransE [3], TransR [4] and TransD [11]. They take relations as translations of the head and tail entity embeddings. Another class of similarity-based HIN embedding methods uses bilinear methods, such as RESCAL [14], ComplEx [13], and DistMult [12]. These methods represent relation embeddings as a transformation of the head and tail entity embeddings [15]. There are also meta-path-based methods [16], and meta-graph-based methods [17], utilizing the structural features in an HIN as attempts to align the path-based or subgraph-based distributions.

Despite their success, these works assume only one source in the HIN and do not consider the distributional difference among sources of subgraphs. And there is a need to align the distributions of feature embeddings from different sources of the HIN to improve downstream task performance. Without loss of generality, we focus on similarity-based embedding methods to illustrate our distribution alignment approach. Our method can be easily applied to all HIN embedding methods on multi-source HINs in general as the alignment is performed on samples of node and relation type embeddings.

Recently there are methods using GNNs to learn the representations of an HIN [7, 9, 18, 19, 20, 21]. Although GNNs can extract the enriched semantic information contained in the HIN, the embeddings of these models are often trained on a supervised or semi-supervised basis with respect to a specific task. Label information on nodes and edges needs to be provided for satisfactory embedding learning. And they can hardly be generalized when the embeddings need to be applied to another task. Additionally, most GNN-based methods work with the adjacency matrix of the HIN, e.g., graph convolutional neural network (GCN) [18] and its variants [1] on HIN perform node aggregation based on the transformed adjacency matrix. These matrices cannot be processed by the memory. Therefore, it is difficult to apply GNN-based HIN embedding methods for large-scale tasks such as recommendation systems which contain networks with billions of user nodes and millions of movies.

In contrast, the aforementioned similarity-based HIN embedding methods perform embedding learning on edge samples, which allows parallelism and therefore scalability. Since the trained embeddings learn HIN representations by encoding the similarity, the similarity features of the HIN are not associated with a specific task. These properties motivate us to propose a multi-source HIN representation learning framework which is not only applicable to any downstream task but also is scalable to large HINs.

*B. Distribution Alignment*

Distribution alignment, also known as domain adaptation in transfer learning, has been a key topic in HIN representation learning, as the heterogeneity in entities and relations introduces misalignments in their respective distributions. There are many attempts in existing work to align the distributions of key features in an HIN. Transformation approaches aim to learn a transformation matrix or an attention mechanism to translate the feature embeddings of different types (nodes or edges) into the same embedding space [7, 8]. Most of the similarity-based methods mentioned above also attempt to align the feature embeddings between entities and relations in an HIN [3, 4, 11, 12]. For example, TransE [3] approximates distribution of the tail node embedding in an edge by the sum of head and relation embeddings. Heterogeneous graph attention network (HAN) [8] adopts a learnable transformation layer to each node type to transform the node embeddings into a space invariant to node types.

Adversarial learning approaches introduce discriminator networks as a domain classifier whose losses are used to measure high-dimensional distribution differences [10, 19, 22, 23]. Moreover, several works applied distance measures such as the maximum mean discrepancy (MMD) to perform distribution alignment [9], these works aim to minimize the distances between distributions to align the distributions of feature embeddings. These alignment methods have been extensively applied to domain adaptation to improve transfer learning performance among multiple graphs. However, these methods are never introduced to align the feature distributions within an HIN.

Inspired by the above works in distribution alignment, we include both the distance measure approach and the adversarial approach in our proposed framework. We use these alignment methods to align the distributions of HIN embeddings with respect to sources, in addition to their original attempts to align the distributions of nodes or edge types. We assess the performance of these distribution alignment methods in aligning the embedding distributions by experiments on different downstream tasks, such as node classification and link prediction.

## III. PROBLEM STATEMENT

### A. Definitions

**Heterogeneous Information Network**: A heterogeneous information network is defined by a graph $\mathcal{G} = (\mathcal{V}, \mathcal{E}, \mathcal{A}, \mathcal{R})$ where $\mathcal{V}, \mathcal{E}, \mathcal{A}, \mathcal{R}$ represent the set of entities (nodes), relations (edges), entity types, and relation types, respectively. A triple in E is defined by e = (h, r, t), where h, t $\in \mathcal{V}$ are the heads and tails nodes representing the entities in $\mathcal{G}$, and $r \in \mathcal{R}$ represents the type of relation connecting the entities. For $v \in \mathcal{V}$, $v$ is mapped to an entity type by a function $\tau(v) \in \mathcal{A}$, and $r$ is mapped to a relation type by a function $\phi(r) \in \mathcal{R}$.

**Heterogeneous Information Network Embeddings**: We encode the similarity of each node in the HIN to a $d$-dimensional vector with a similarity function $f(e)$. The node and edge type embeddings can be used as input features for training an arbitrary downstream task model.

### B. Problem: Multi-Source Heterogeneous Information Network Embeddings

Consider a heterogeneous information network $\mathcal{G} = (\mathcal{V}, \mathcal{E}, \mathcal{A}, \mathcal{R})$, let $\mathcal{S}$ represents the set of sources in $\mathcal{G}$. We have a series of $K = |\mathcal{S}|$ subgraphs $\{\mathcal{G}_i\}_{i=1}^K = \{(\mathcal{V}_I, \mathcal{E}_I, \mathcal{A}_I, \mathcal{R}_i)\}_{i=1}^K$ as the predefined sources of $\mathcal{G}$. Let $\mathcal{X}$ be the embeddings space of nodes and edge types in $\mathcal{G}$, and let $\mathcal{X}_i$ denote the embedding space of nodes and edge types in each subgraph $\mathcal{G}_i$.

We wish to assign an embedding x ∈ X to each node and edge type in $\mathcal{G}$. We also wish to align the distributions of $\{\mathcal{X}_i\}_{i=1}^K$ such that for a model $\mathcal{M}$ trained on graph $\mathcal{G}$, on a given downstream task $\mathcal{T}$, the model $\mathcal{M}$ can perform accurately.

## IV. Methodology

We introduce SUMSHINE in this section. The major component of SUMSHINE consists of a source-aware negative sampling strategy and a loss function designed to regularize distribution discrepancies across subgraphs. Conceptual visualization of the training paradigm of SUMSHINE is shown in Figure 1.

### A. Source-Aware Negative Sampling

Given a positive edge e = (h, r, t), negative sampling replaces either a head or a tail (but not both) by another arbitrary node in the HIN to produce negative edges which do not exist in the original graph [3, 5]. The embeddings can be learned by maximizing the distance between the positive samples (i.e., ground truth edges) and the negative samples. However, sampling from imbalanced subgraphs leads to data imbalance problem between subgraph sources. Edges in larger subgraphs (such as a user interaction graph) are sampled more often than the smaller subgraphs (such as an album knowledge graph). To rebalance the data with respect to sources, we introduce source-aware negative sampling to sample edges uniformly from each subgraph source. By source-aware sampling we can balance the number of edges sampled by sources, and reduce the bias on embeddings from data imbalance. For each subgraph source $\mathcal{G}_i$ in $\mathcal{G}$, we sample a fixed-size batch of edges from it to match the dimensions of sample embedding matrices. Given and edge $e_i = (h_i, r_i, t_i)$ from a source $\mathcal{G}_i$, we select a set of negative samples $S'_{e_i}$ by replacing either a head node by $h'_i$ or a tail node by $t'_i$, where $h'_i$ and $t'_i$ are entities other than $h'_i$ or $t'_i$ within the subgraph. We denote the set of negative samples as

$$S'_{e_i} = \{(h'_i, r_i, t_i) | h'_i \in \mathcal{V}_i\} \cup \{(h_i, r_i, t'_i) | t'_i \in \mathcal{V}_i\}.$$

The negative samples are combined with a batch of positive edges to compute the similarity on a mini-batch basis. The similarity-based loss function is given by

$$\mathcal{L}_{sim} = \sum_{i=1}^{K} \sum_{e_i \in \mathcal{G}_i} \sum_{e'_i \in S'_{e_i}} [f(e_i) - f(e'_i) + \gamma]_+,$$

where $\gamma$ is the margin and $[x]_+ = \max(x, 0)$. The scoring function $f(e)$ is uniquely defined by the HINE method. We assume the embeddings of the edge samples are independent and identically distributed (IID). We use mini-batch gradient descent [5] to back-propagate the similarity loss to the embeddings to learn the HIN representation.

### B. Aligning Sources with Regularization

As mentioned above, one of the key issues we want to ad-dress here is to alleviate the distribution discrepancies among different subgraphs. More specifically, given an arbitrary pair of subgraphs in $\{\mathcal{G}_i\}_{i=1}^K$, we define the distribution functions $P$ and $Q$ on the embedding space to be the embedding distributions on the two subgraphs, and we aim to encourage less distribution discrepancy between $P$ and $Q$ despite their domain differences. To achieve this, we introduce two regularization methods — distance-measure-based regularization and adversarial regularization.

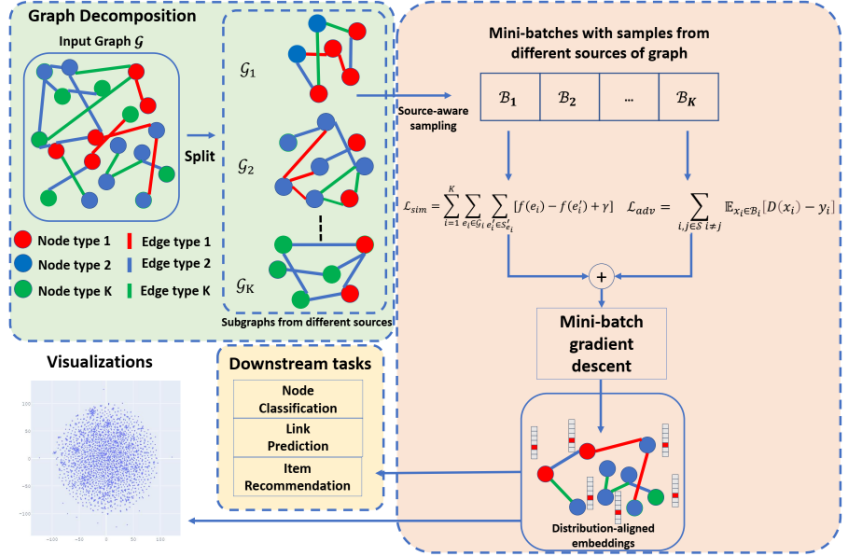

Fig. 1: Training paradigm of our proposed SUMSHINE HIN Embedding method.

We first introduce distance-measure-based regularization. In this paper, we adopt the distance measures MMD [24], the Kullback–Leiber (KL) divergence, and the Jensen–Shannon (JS) divergence [25] in our experiments, while our framework can be generalized to incorporate any distance measures. We use $\Delta$ to denote the distribution distance between $P$ and $Q$. The KL divergence on $P$ and $Q$ is defined as

$$\Delta_{KL}(P \parallel Q) = \sum_{x \in \mathcal{X}} P(x) \log\left(\frac{P(x)}{Q(x)}\right),$$

and the JS divergence is the symmetric and smoothed version of the KL divergence defined by

$$\Delta_{JS}(P \parallel Q) = \frac{1}{2}\left(\Delta_{KL}(P \parallel Q) + \Delta_{KL}(Q \parallel P)\right).$$

The MMD loss is a widely used approach to alleviate the marginal distribution disparity [26]. Given a reproducing kernel Hilbert space (RKHS) $\mathcal{H}$ [24], MMD is a distance measure between $P$ and $Q$ which is defined as

$$MMD(P \parallel Q) = \left\| \mu_P - \mu_Q \right\|_{\mathcal{H}},$$

where $\mu_P$ and $\mu_Q$ are respectively the kernel means computed on $P$ and $Q$ by a kernel function $k(\cdot)$ (e.g., a Gaussian kernel). We perform distribution alignment between pairs of subgraphs. For each batch sampled by source-aware sampling and each pair of sources, we compute the distribution differences of embeddings for both relation types and entities, using one of the distance measures introduced above. The regularization loss $\mathcal{L}_{dist}$ is the sum of distribution distances for both entity and relation type embeddings over each pair of sources. The total loss can be obtained by combining $\mathcal{L}_{dist}$ with the similarity loss to propagate both the similarity and the distribution discrepancy into HIN embedding training,

$$\mathcal{L}_{tot} = \mathcal{L}_{dist} + \lambda \mathcal{L}_{sim} \quad (1)$$

where $\lambda$ is a tuning parameter.

Alignment methods based on distance measures heavily relies on the measure chosen, and the high dimensional distribution difference such as the geodesic difference may not be incorporated by the measure. R Connor et. al [27] suggested that the high dimensionality of data in the metric space may cause metrics of distribution differences to be biased. Adversarial Regularization, on the contrary, uses a feedforward network as a discriminative classifier to capture the distributional differences in high dimension to avoid comparing high-dimensional data in the metric space directly and ameliorate the bias compared to the aforementioned distance measures [28].

With the recent development of adversarial distribution alignment [10, 23, 28, 29], we introduce adversarial regularization to HIN embedding training. We consider the embeddings from different subgraphs trained by an HIN embedding method as the generated information, and use an adversarial discriminator D as a domain classifier to classify the source of the embeddings. As a result, we consider the loss from the discriminator a measure of distribution discrepancy between the sources and use it to align the embeddings distributions from different sources [10, 28].

Let $\mathcal{B}_i \subseteq \mathcal{X}_i$ be the node and edge type embeddings in a sampled batch from a subgraph source $\mathcal{G}_i$. The discriminator receives the batch of embeddings $\mathcal{B}_i$, and generates the probability that which source these embeddings are from. The predictions are compared with the ground truth one-hot label $y_i$, where its i-th entry is 1 with the rest being zeros. The loss of the discriminator is given by

$$\mathcal{L}_D = \sum_{i \in \mathcal{S}} \mathbb{E}_{x_i \in \mathcal{B}_i}[D(x_i) - y_i].$$

We then compute the adversarial loss and combine it together with the similarity loss. We compute the distribution distance by inverting the true label $y_i$ to $y_j$ where $i \neq j$. The adversarial loss is then given by

$$\mathcal{L}_{adv} = \sum_{i \in \mathcal{S}} \mathbb{E}_{x_i \in \mathcal{B}_i} [D(x_i) - y_j].$$

The loss $D(x_i) - y_j$ for each pair of sources $i$ and $j$ indicates the distributional difference between them. We aim to include this adversarial loss to the embeddings such that the embeddings can be more similar in distribution to fool the discriminator. We then multiply the adversarial loss by the tuning parameter $\lambda$ and compute the aggregated loss using equation (1) with $\mathcal{L}_{dist}$ replaced by $\mathcal{L}_{adv}$.

## V. THEORETICAL ANALYSIS

We provide theoretical analysis to show why aligning the distribution of embeddings from different sources of subgraphs in a heterogeneous graph can improve the downstream task performance, where the error of generalization will be bounded by probability with an optimized bound.

**Settings** We first define the loss of generalization. When generalizing the model from a origin environment to the target environment on the same task $\mathcal{T}$, we want to minimize the generalization bound $\epsilon$ such that the error of generalization is bounded by $\epsilon$ in probability, which is for any $\delta > 0$,

$$\mathbb{P}(|\mathcal{L}_{org} - \mathcal{L}_{dest}| \leq \epsilon) \leq 1 - \delta,$$

where $\mathcal{L}_{org}, \mathcal{L}_{dest}$ are the losses of the origin and destination respectively for the downstream task $\mathcal{T}$ and $|\mathcal{L}_{org} - \mathcal{L}_{dest}|$ is the error of transferal. We further assume that the source discrepancy leads to the largest generalization error than any pairs of subgraphs in $\mathcal{G}$, which is formulated in assumption 1.

**Assumption 1** Suppose $\{\mathcal{G}\}_{i=1}^{K}$ is the set of $K$ pre-defined subgraph sources of $\mathcal{G}$, let $\mathcal{G}_{s_1^*}, \mathcal{G}_{s_2^*}$ be the pair of subgraphs in $\{\mathcal{G}\}_{i=1}^{K}$ which has the largest generalization error,

$$\left|\mathcal{L}_{\mathcal{G}_{s_1^*}} - \mathcal{L}_{\mathcal{G}_{s_2^*}}\right| \geq \left|\mathcal{L}_{\mathcal{G}_{s_1}} - \mathcal{L}_{\mathcal{G}_{s_2}}\right| \forall s_1, s_2 \in \mathcal{S},$$

where $\mathcal{L}_g$ is the downstream task loss using a graph $g$, then we assume that for any pair of subgraph in $\mathcal{G}$, the generalization loss is less than or equal to $\left|\mathcal{L}_{\mathcal{G}_{s_1^*}} - \mathcal{L}_{\mathcal{G}_{s_2^*}}\right|$. This assumption is reasonable since the sources of subgraphs are mostly having the largest semantic difference and least overlaps. We can focus on minimizing the source-level embedding distribution discrepancy with this assumption.

To obtain a theoretical bound of source-level generalization error, we generalize the current pairwise analysis from Zhang *et al.* [10] to multiple sources. Given a specific downstream $\mathcal{T}$ and a series of true labels $\{\hat{P}_i(y|z)\}_{i=1}^{K}$ for each source $i$ in $\mathcal{S}$. $\mathcal{L}_i$ is the downstream task loss for source $i$, and $p_i(z)$ is the density function of a given node from source $i$ with embedding $z$ in the shared-semantic embedding space $Z$ and $\mathcal{M}_i$ be the downstream task model trained to make prediction $\hat{y} = \mathcal{M}_i(z)$ [10],

$$\mathcal{L}_i = \int_Z p_i(z) \Delta(\hat{P}_i(y|z), P(y|z)) dz,$$

where $\Delta$ is the divergence function to determine the loss of predicted labels to ground truth labels. We have the following theorem:

**Theorem 1** If the following conditions are satisfied:

$$|\hat{P}_i(y|z) - \hat{P}_j(y|z)| < c_{ij}, \forall i,j \in \mathcal{S}$$

$$\frac{|p_i(z) - p_j(z)|}{p_i(z)} < \epsilon_{ij}, \forall i,j \in \mathcal{S}$$

Then we have

$$\sum_{i=1}^{K} \sum_{i \neq j}^{K} \sum_{j=1}^{K} (\mathcal{L}_i - \mathcal{L}_j) \leq \sum_{i=1}^{K} \sum_{i \neq j}^{K} \sum_{j=1}^{K} (\mathcal{L}_i \epsilon_{ij} - c_{ij} + c_{ij} + \epsilon_{ij}).$$

Theorem 1 states that if we want to control the generalization loss from each source $i$ to any other sources, we need to align both the semantic meaning and the distributions $p_i(z)$ of the embeddings by controlling every pairwise distance. The proof of theorem 1 is given by Zhang et al. [10]. Since all the subgraphs are trained jointly and the subgraph embeddings are essentially having the same semantic meaning, we further assume $c_{ij} \to 0$ in theorem 1 as the embeddings having very close semantic meanings (i.e., ground truth labels will be the same for a given $z$). Then we have the following corollary:

**Corollary 1** If $c_{ij} \to 0$, we have the following reduced version of theorem 1:

$$\sum_{i=1}^{K} \sum_{i \neq j}^{K} \sum_{j=1}^{K} (\mathcal{L}_i - \mathcal{L}_j) \leq \sum_{i=1}^{K} \sum_{i \neq j}^{K} \sum_{j=1}^{K} \mathcal{L}_i \epsilon_{ij}. \quad (2)$$

Equation (2) indicates that in order to reduce the generalization error between any pairwise environments, we only need to minimize the distribution difference of all pairs of environments. In other words, we want to minimize which can be achieved by minimizing the adversarial loss $\mathcal{L}_{adv}$. On the other hand, the similarity loss $\mathcal{L}_{sim}$ can highlight the node and edge features in the graph, thus $\mathcal{L}_i$ can still be minimized.

## VI. THEORETICAL ANALYSIS

### A. Datasets

We collect public datasets for benchmarking HIN embedding methods that contain multiple sources: WordNet18 (WN18) [3] and DBPedia (DBP). Table II provides a summary of the datasets used for experiments. We also compose a real

TABLE I: Example heterogeneous graph embedding methods and their scoring functions

| Model | Embedding Space | Relation Embeddings | Scoring Function | Space Complexity |
|---|---|---|---|---|
| TransE [3] | $h, t \in R^d$ | $r \in R^d$ | $\|h + r - t\|_p$ | $O(d)$ |
| TransR [4] | $h, t \in R^d$ | $r \in R^d, M_r t \in R^{k \times d}$ | $\|M_r h + r - M_r t\|_p$ | $O(d^2)$ |
| TransD [11] | $h, t, M_h, M_t \in R^d$ | $r, M_r \in R^d$ | $-\|(M_r M_h^T + I)h + r - (M_r M_h^T + I)t\|_p$ | $O(d^2)$ |
| RESCAL [14] | $h, t \in R^d$ | $M_r t \in R^{d \times d}$ | $h^T M_r t$ | $O(d^2)$ |
| DistMult [12] | $h, t \in R^d$ | $r \in R^d$ | $h^T diag(r) t$ | $O(d)$ |
| ComplEx [13] | $h, t \in C^d$ | $r \in C^d$ | $h^T Re(diag(r)) t$ | $O(d)$ |

$h, r, t$: embeddings of head, relation, and tail; $d$: dimension of the embedding vector; $Re(z)$: real part of complex number $z$; $diag(x)$: diagonal entries of matrix $x$; $C^d$: complex space of dimension $d$; $M_r, M_t$: learnable matrices to transform the relation or tail embeddings.

TABLE II: Datasets Summary

| Dataset | $|\mathcal{V}|$ | $|\mathcal{R}|$ | $|\mathcal{E}|$ | $|\mathcal{A}|$ |
|---|---|---|---|---|
| DBP-Total[1] | 118,907 | 305 | 118,907 | 1 |
| DBP-WD[2] | 42,201 | 259 | 60,000 | 1 |
| DBP-YG[3] | 37,805 | 236 | 60,000 | 1 |
| MRec-Total | 284,908 | 8 | 307,029 | 1 |
| MRec-Album | 57,203 | 5 | 62,915 | 1 |
| MRec-User | 235,693 | 3 | 246,629 | 1 |
| WN18-Total | 40,943 | 18 | 151,442 | 1 |
| WN18-A | 39,398 | 6 | 96,598 | 1 |
| WN18-B | 20,179 | 7 | 41,836 | 1 |
| WN18-C | 7,516 | 5 | 13,008 | 1 |

[1] Total: The whole graph constructed by merging the subgraph sources
[2] WD: Wikidata source of DBPedia
[3] YG: WordNet source of DBPedia

dataset, namely MRec (Movie Recommendation), based on real user movie watching data from a practical recommendation system. MRec has two sources — one is representing the user-movie interaction graph, containing the users' movie-watch histories, and another one is simulating the knowledge graph of the album of movies with ground truth entities related to the movies such as tags, directors, and actors. We use the MRec dataset to model the distribution difference caused by graph sizes in HINs. To validate the performance when our method is applied to more than two sources, we perform experiments on the WN18 dataset which contains three sources of subgraphs — namely A, B, and C. The subgraphs are created by categorizing the relations according to their semantic meanings so that different subgraphs will correspond to different sets of relations, incurring different average node degree per relation type. Details on the sources can be found in the Appendix.

For node classification, we collect channel labels from the MRec dataset for 7000 movie nodes where these nodes are present in both the user interaction graph and album knowledge graph. Each movie node is labelled by one of the following six classes: "not movie", channel 1 to 4, or "other movie" (i.e. channel information not available). We additionally sample 3000 "not movie" (i.e., negative) entities from the MRec data for training in order to produce class-wise balanced data. We randomly choose 7000 movie entities and 3000 non-movie entities from the testing graph as the testing data.

### A. Benchmarking Methods

We compare our method against the baseline HIN embedding learning methods, including TransE [3], TransR [4], and DistMult [12], and validate the improvements provided by our method. We also show the performance of GNN-based approaches [18, 20, 21], of which the main goal is to learn node embeddings for a specific downstream task, as a reference. For simplicity, we use the scoring function of TransE [3] in our proposed framework, while the performance of our method with other scoring functions is presented by ablation studies in Section VII-A. To validate the effectiveness of our approach, we apply the node and edge type embeddings produced by each approach as the feature input to downstream tasks. Table I presents a summary of the embedding methods and their scoring functions

Descriptions of each method are listed below:

- **TransE** [3]: Learning the relations in a multi-relation graph by translating the source and destination node embeddings of the relation.
- **TransD** [11]: In addition to TransR translating the relation space, TransD also maps the entity space to a common latent space.
- **TransR** [4]: Building entities and relations in separate embedding spaces, and project entities to relation space then building translation between the projected entities.
- **RESCAL** [14]: RESCAL is a bilinear model that captures latent semantics of a knowledge graph through associate entities with vectors and represents each relation as a matrix that models pairwise interaction between entities. Entities and relations are represented as a multi-dimensional tensor to factorize the feature vectors to rank $r$.
- **DistMult** [12]: Improving the time complexity of RESCAL to linear time by restricting the relationship to only symmetric relations.

### B. Experiment Settings

We perform inductive link prediction [30] as the downstream task to validate our framework. After we obtain the node and edge type embeddings produced by different HIN embedding approaches, we use a multiple layer perceptron (MLP) matcher model to perform the downstream task. A matcher model is a binary classifier that output the probability of having a link given the edge embedding (i.e., concatenated embedding of head, tail and relation) as the input. For GNN baselines, we directly train a GNN to perform link prediction instead of MLP. A matcher model can perform inductive link prediction across subgraphs rather than transductive [30] link prediction which can only predict linkage with the observed data (i.e., the subgraph used for training).

To highlight the advantage of combining subgraphs and learning embeddings in distribution-aligned latent embedding space, we design an experiment setting for inductive link prediction as follows: When training for the downstream tasks, we only take the training data that contain edges from one subgraph while keeping the data which contain edges from other graphs as evaluation data. Note that we borrow this setting from the literature on GNN transfer learning [9, 10], where the goal of these works is to transfer the GNN model from one graph to another. However in our setting, rather than showing how transferrable the downstream task models are, we show how a distribution-aligned embedding training mechanism can benefit the downstream task performance, especially when there are distribution shifts among subgraphs. For demonstration of results, we denote the training-testing split in each link prediction experiment with an arrow "Training $\rightarrow$ Testing" for notational convenience.

For each testing edge, we replace the head and then the tail to each of the 1000 negative entities sampled from the testing entities. We rank the true edge together with its negative samples according to the probability that an edge exists between the head and tail output from the MLP matcher model. We sample 1000 negative entities to corrupt the ground truth edge instead of all the testing entities in the subgraph because scaling the metrics can enhance the comparability among datasets. Since each testing entity has an equal probability to be replaced, the downstream task performance is not affected by the choice of the sample size of negatives.

We use node classification as another downstream task. We first train an MLP node classification model on one source of subgraph and then test the model on another source. The classification model takes an HIN node embedding as the input, and classify the node to one of the six classes according to its embedding.

We evaluate the link prediction performance using Hits@$n$ and mean reciprocal rank (MRR) and the node classification performance using classification accuracy. More details on evaluation metrics and model configurations are presented in the appendix.

### C. Link Prediction

We validate our framework by inductive link prediction. Table III provides a summary of the prediction performance of our method to various baselines. We choose the JS divergence to be the distance measure for alignment. More discussions on the effects of different distance measures will be included in section VII-B. The experiments are performed on MRec and DBPedia datasets with two sources. We observe that the link prediction results after distribution alignment, with either adversarial regularization or distance-measure-based regularization, are uniformly better among the benchmarks for all evaluation metrics. The performance of adversarial regularization is superior to the JS divergence, which supports the superiority of adversarial alignment over distance-measured-based alignment. The results show that inductive link prediction is optimized for multi-source graphs if we align the distributions of the embeddings.

We also observe that GNN models underperform our method in most of the inductive link prediction tasks. GNN link prediction models can extract global features by aggregating node features from the whole graph (e.g., through the transformed adjacency matrix), which is more capable than similarity-based method focusing on local similarity features. However, the misalignment in subgraph sources still decrease the performance of GNN-based link prediction models, which make them underperform our model in general. Additionally, out-of-memory errors were reported for GNN models when the size of user graph is doubled in the User$\rightarrow$Album experiment. This highlights the scalability constraints of GNN models.

We further validate our framework on datasets with more than two sources. Table IV presents the results of inductive link prediction performance for each of the six training-testing splits on the WN18 dataset. We observe that in most of the tasks the performances are improved with distribution aligned embeddings. This validate the consistency of our framework when $K$ is generalized to be larger than 2 (i.e., multiple sources).

The MRec dataset is simulated to have a significant imbalance of data with respect to sources. Hence the data without source-aware sampling are mostly sampled from the user interaction graph and only a few of them are from the album knowledge graph. It is noteworthy that since the user-interaction graph is sparse (i.e. as users have diverged interests), the link prediction model trained on the album knowledge graph, is heavily biased and less transferrable to the user-interaction graph, leading to a performance which occasionally worse than a random guess.

With source-aware sampling, smaller subgraphs can be sampled equal times to larger subgraphs. Therefore, the information in the smaller subgraphs can be leveraged especially when there is a large degree of data imbalance among the sub- graphs. Hence source-aware sampling significantly increases the awareness to small subgraphs, which resolves the data imbalance problem in existing methods.

### D. Node Classification

Table V presents the node classification performance with or without distribution alignment respectively. We observe that there are improvements in accuracy for both user to album and album to user transferal tasks. Note that the MRec dataset contains subgraphs with significant different average node degrees. Therefore, without taking into account the imbalance issue, the node and edge type embeddings will be dominated by the semantic information contained in the user interaction graph. With the help of distribution alignment during embedding training, the structure information in the movie-knowledge graph can be leveraged and ameliorate the domination of the user-interaction graph, hence the recall may be higher while the precision is sacrificed to adjust the bias caused by a large difference in average node degrees.

TABLE III: Link prediction performance of SUMSHINE to baseline methods on DBPedia and MRec datasets (JS: Regularization loss is the JS Divergence; ADV: Regularization loss is the adversarial loss).

| Model | DBP | | | | MRec | | | |
| --- | --- | --- | --- | --- | --- | --- | --- | --- |
| | WD→YG | | YG→WD | | User→Album | | Album→User | |
| | MRR ↑ | Hit@10 ↑ | MRR | Hit@10 | MRR | Hit@10 | MRR | Hit@10 |
| TransE [3] | 0.0130 | 0.0232 | 0.0117 | 0.0232 | 0.0060 | 0.0055 | 0.0059 | 0.0067 |
| TransR [4] | 0.0295 | 0.0638 | 0.0302 | 0.0632 | 0.0670 | 0.1100 | 0.0051 | 0.0044 |
| GIN [20] | 0.0293 | 0.0607 | 0.0259 | 0.0498 | 0.0275 | 0.0559 | 0.0587 | 0.1290 |
| GCN [18] | 0.0276 | 0.0511 | 0.0244 | 0.0435 | **0.1337** | **0.1643** | 0.0608 | 0.0981 |
| GAT [21] | 0.0368 | 0.0653 | 0.0288 | 0.0553 | 0.0414 | 0.0834 | 0.0305 | 0.0674 |
| SUMSHINE-JS | 0.0474 | 0.1236 | 0.0320 | 0.0653 | 0.0149 | 0.0262 | 0.0380 | 0.1040 |
| SUMSHINE-ADV | **0.0487** | **0.1257** | **0.0536** | **0.1022** | 0.1232 | 0.1549 | **0.1954** | **0.3386** |

TABLE IV: Link prediction performance of our method to TransE on the WordNet18 dataset which has three sources. The similarity loss used for SUMSHINE is the same as TransE.

| | TransE | | SUMSHINE | |
| --- | --- | --- | --- | --- |
| Data Split | MRR ↑ | Hit@10 ↑ | MRR | Hit@10 |
| A → B | **0.0073** | 0.0089 | 0.0072 | 0.0104 |
| B → A | 0.0069 | 0.0079 | **0.0080** | 0.0126 |
| B → C | 0.0064 | 0.0103 | **0.0087** | 0.0115 |
| C → B | 0.0061 | 0.0100 | **0.0075** | 0.0136 |
| A → C | 0.0067 | 0.0092 | **0.0099** | 0.0138 |
| C → A | 0.0081 | 0.0134 | **0.0086** | 0.0123 |

TABLE V: Node classification performance (in classification accuracy) of TransE with or without distribution alignment respectively.

| Data Split | TransE | SUMSHINE |
| --- | --- | --- |
| User → Album | 0.5392 | 0.5548 |
| Album → User | 0.5497 | 0.6249 |

*E. Visualization*

To validate the performance of our alignment method, we use Isomap plots to visualize the trained embeddings with or without distribution alignment, respectively. The high-dimensional information such as geodesic distance can be preserved by Isomap when reducing the dimension of the embedding distribution. Figure 2 shows the Isomap plot of the embedding trained by TransE and SUMSHINE on the DBPedia dataset and the MRec dataset. More visualizations are shown in the appendix.

It is observed that with distribution alignment, the distributions of embeddings in YG and WD are smoother (i.e. having fewer random clusters and more flat regions), while the source-invariant features such as modes of distributions are still preserved by similarity learning. The alignment in distributions can also be validated quantitatively by computing the JS divergences without and with adversarial regularization respectively, which is shown in Table VI. We observe that the distribution discrepancy is decreased significantly after adversarial alignment. According to the flat-minima hypothesis [31], smooth regions are the key for smooth transferal of the features between distributions, which allow better alignments in features from the subgraphs. The downstream task models can hence make use of the aligned features to improve their performances.

TABLE VI: JS Divergences of the trained embeddings of DBP and MRec with respect to their sources (User and Album for MRec; WD and YG for DBP). The comparison is performed between distribution-aligned embeddings (SUMSHINE) and the original embeddings (TransE).

| Data Split | TransE | SUMSHINE |
| --- | --- | --- |
| User — Album | 8.7493 | 0.0831 |
| WD — YG | 0.5870 | 0.1522 |

ABLATION STUDIES

*A. Impact of Scoring Functions*

We experiment with other HIN embedding methods by exploring different scoring functions. Table VII demonstrates the performance on link prediction using the embeddings with or without distribution alignment respectively. Similar to that of TransE, we observe that distribution alignment can still improve inference performance when the scoring function is altered. We can verify that the performance of our framework is invariant to the changes in scoring functions, which indicates that by training distribution-aligned HIN embeddings the downstream tasks can perform more accurately with any chosen scoring function. This ensures the extensibility of our framework when new HIN embedding methods are developed.

*B. Impact of Distance Measures*

We further evaluate the performance of our model when the distance measure is changed to another one, e.g., the KL divergence or MMD. Table VIII presents the performance of our framework on link prediction on the DBPedia dataset when using different distance measures. We observe that both distance measures can align the distributions of embeddings and improve the downstream task performance. Since MMD computes the distribution distance in a Hilbert space [24], it can incorporate higher-dimensional features than the KL divergence, the link prediction performance with MMD is better than that with the KL divergence, while the time complexity of MMD is higher. We conclude that using distance measures can align the distributions and improve the embedding quality, with small variations to the distance measure selected.

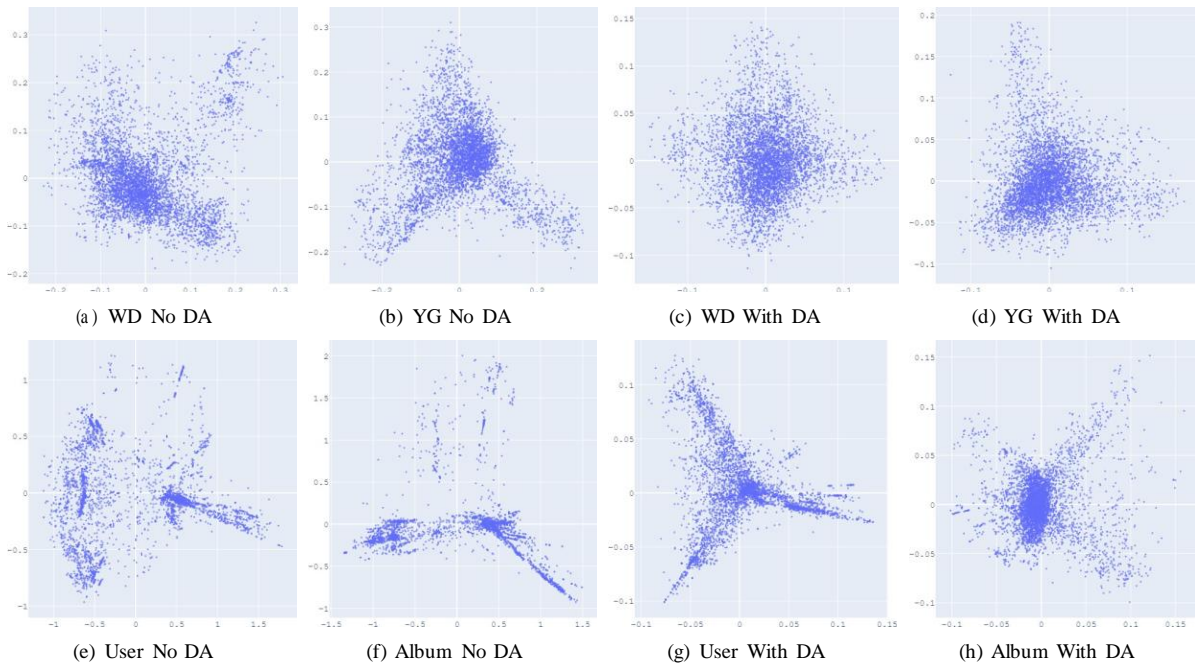

(a) WD No DA  (b) YG No DA  (c) WD With DA  (d) YG With DA

(e) User No DA  (f) Album No DA  (g) User With DA  (h) Album With DA

Fig. 2: Isomap plots of the embeddings of DBP by sources WD and YG, and MRec by sources User and Album, with or without distribution alignment (DA) respectively. The alignment method used is adversarial regularization.

TABLE VII: Link prediction performances of different similarity functions. The alignment method is adversarial regularization.

| Model | User → Album MRR | Hit@3 | Album → User MRR | Hit@3 |
|---|---|---|---|---|
| TransE | 0.0064 | 0.0024 | 0.0084 | 0.0030 |
| SUMSHINE-TransE | 0.1232 | 0.1421 | 0.1954 | 0.2358 |
| TransR | 0.0670 | 0.0758 | 0.0051 | 0.0020 |
| SUMSHINE-TransR | 0.1904 | 0.2297 | 0.2208 | 0.2850 |
| TransD | 0.0059 | 0.0023 | 0.0088 | 0.0047 |
| SUMSHINE-TransD | 0.1381 | 0.1703 | 0.1144 | 0.1567 |
| DistMult | 0.0206 | 0.0169 | 0.0448 | 0.0555 |
| SUMSHINE-DistMult | 0.0238 | 0.0245 | 0.0688 | 0.0704 |
| RESCAL | 0.0053 | 0.0022 | 0.0326 | 0.0350 |
| SUMSHINE-RESCAL | 0.0428 | 0.0495 | 0.1109 | 0.1300 |

TABLE VIII: Link prediction performances of alignment using different distance measures on the DBPedia dataset. The similarity loss for SUMSHINE the same as TransE.

| Model | WD → YG MRR | Hit@3 | YG → WD MRR | Hit@3 |
|---|---|---|---|---|
| TransE | 0.0130 | 0.0117 | 0.0064 | 0.0056 |
| SUMSHINE-KL | 0.0283 | 0.0221 | 0.0251 | 0.0235 |
| SUMSHINE-MMD | 0.0471 | **0.0534** | 0.0280 | 0.0204 |
| SUMSHINE-JS | **0.0474** | 0.0518 | **0.0320** | **0.0330** |

### C. Impact of Subgraph Sizes

The difference in sizes among the subgraphs highlights the distribution discrepancies. The aforementioned size difference between the user-interaction graph and album knowledge graph is a typical example of the size difference. We further study how SUMSHINE performs as the ratio of sizes (in the number of edges) changes. We compose different variants of the MRec dataset with different ratios of the total number of edges — from an approximately equal number of edges to large differences in the total number of edges. We compare link prediction performance with the original TransE and the distribution-aligned version with adversarial regularization and use MRR as the evaluation metric.

Figure 3 demonstrates a decreasing trend of MRR as the ratio (album: user) of the number of edges changes from 1:4 to 1:1, which indicates that our framework has better performance when the size between subgraphs has larger differences. On the other hand, TransE is having improving performance as the number of edges of the subgraphs are close to each other. However, the link prediction performance of TransE is still lower without distribution alignment. The reason is that the user interaction graph has diverged features where the features cannot be smoothly transferred without distribution alignment. For application on graph-based recommendation systems where the user interaction graph and album graph typically have a large difference in graph size, our framework performs better to resolve the information misalignment problem for better recommendation performance. This is a practical

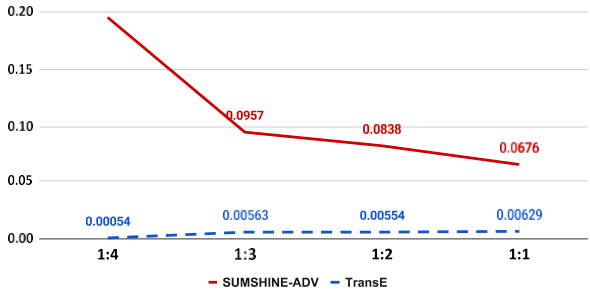

Fig. 3: Performance (in MRR) of album→user link prediction task of SUMSHINE-ADV with respect to different ratios of the number of edges (A:U) between the album knowledge graph (A) and the user interaction graph (U).

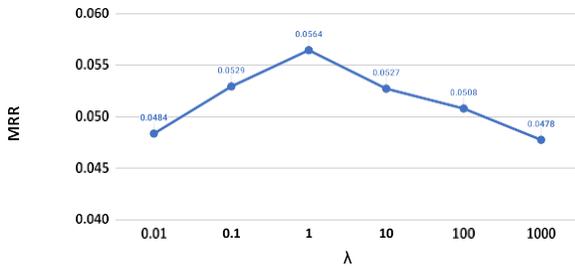

Fig. 4: Link prediction performance (in MRR) of SUMSHINE-ADV on DBPedia dataset with respect to different values of $\lambda$. Here the YG source is the training set and the WD source is the testing set.

insight of the above results on the industrial application of our framework.

### D. Impact of Tuning Parameter $\lambda$

We study the impact of tuning parameter $\lambda$ in equation (1) on the performance of our method. We explore a grid of values of $\lambda$: [0.01, 0.1, 1, 10, 100, 1000] and perform adversarial distribution alignment with each $\lambda$ value. Figure 4 shows how our methods perform on DBPedia datasets with different values of $\lambda$. We observe that the optimal performance is obtained when $\lambda$ is 1. We also observe that the link prediction performance is worse when $\lambda$ is too small or too large. When $\lambda$ is too large, the regularization on the embeddings is too heavy such that the similarity feature is not preserved by the embeddings, and the lack of similarity features will decrease the link prediction performance. On the other hand, when $\lambda$ is too small, the misalignment in distribution is not penalized by the alignment loss, the distribution misalignment will also decrease the link prediction performance. Hence $\lambda$ should carefully be tuned to achieve optimal downstream task performance.

### Conclusion and Future Work

We propose SUMSHINE — a scalable unsupervised multi-source graph embedding framework on HINs, which is shown to improve the downstream task performance on the HIN. Extensive experiments have been performed on real datasets and different downstream tasks. Our results demonstrate that the embedding distributions in the subgraph sources of the HIN can be successfully aligned by our method. We also show by ablation studies that the our framework is robust when the distance measure or the scoring function is altered. Additionally, we show that our framework performs better when the sources are having a larger difference in the graph size.

Our framework can be further generalized to integrate multimodal HIN embeddings by aligning the distributions of the side information embeddings such as image or text embeddings. Incorporating multimodality opens the possibility of practical application of our framework to common-sense knowledge graphs where the graph is constructed by merging numerous knowledge bases, including text and image features.

### Acknowledgment

We thank the anonymous reviewers and Dr. Pan Yi Teng for their insights and advice on this research. This work was partially supported by the Research Grants Council of Hong Kong (17308321) and the HKU-TCL Joint Research Center for Artificial Intelligence sponsored by TCL Corporate Research (Hong Kong).

APPENDIX

A. *Relation Decomposition of WN18*

The sources A, B, and C of WN18 are decomposed by relations according to their semantic meaning, where the names of the relations for each source can be found below:

- **A**: instance h̲yponym, hyponym, hypernym, member hol̲onym, instance h̲ypernym, member me̲ronym

- **B**: member_of_domain_topic, synset_domain_usage_of, synset_domain_region_of, member_of_domain_region, derivationally_related_form, member_of_domain_usage, synset_domain_topic_of
- **C**: part_of, verb_group, similar_to, also_see, has_part

### B. Model Configurations

We use *adagrad* as the optimizer with a learning rate of 0.005 and a weight decay of 0.001 for all models. Each positive edge is trained with four negative edges to compute the margin-based loss. The size of the minibatch is 1024. The embeddings of each experiment are trained with 2000 epochs and all the methods converged at this level.

For link prediction, we train a matcher model for each experiment to be an MLP with two hidden layers of hidden dimension 200. The matcher model takes the concatenated head, relation and tail embeddings as the input and has output the softmax probability of having a link. For GNN matcher models (GCN/GAT/GIN), the number of layers is set to be 2 with the final dropout ratio to be 0.4. We train each of the matcher models for 200 epochs in each experiment.

For node classification, we train an MLP classifier with 1 hidden layer of hidden dimension 200 and a softmax output layer for the probability of the six classes. We train the classifier in each experiment for 200 epochs.

### C. Implementation Details

We implement our methods in *Python*. We utilize OpenKE [32] as the backend for loading triples to training and performing link prediction evaluation using the trained embeddings. We also use the *dgl* library [33] to perform graph-related computations and *PyTorch* to perform neural network computations. The models are trained on a server equipped with four NVIDIA TESLA V100 GPUs. The codes and data for the paper are available and will be made public after this paper is published.

### D. Metrics

- Link prediction metrics:
  - **Mean reciprocal rank** (MRR): Mean of reciprocal ranks of first relevant edge. Given a series of query testing edges $Q$, and $\text{rank}_i$ be the rank of a true edge over 1000 negative entities chosen, the MRR is computed by

  $$\text{MRR} = \frac{1}{|Q|} \sum_{i=1}^{|Q|} \frac{1}{\text{rank}_i}$$

  - **Mean rank** (MR): Mean rank of the first relevant edge, subject to larger variance as the high-rank edges which contain diverged features dominate the mean of ranks. MR is computed by

  $$\text{MR} = \frac{1}{|Q|} \sum_{i=1}^{|Q|} \text{rank}_i$$

  - **Hit rate @ $n$**: the fraction of positives that rank in the top $n$ rankings among their negative samples.
- Classification metrics:
  - **Accuracy**: The fraction of correct predictions to the total number of ground truth labels.

### E. Additional Visualizations

Figure 5 presents the visualization results of the embeddings of entities from WN18 dataset, with and without distribution alignment, respectively.

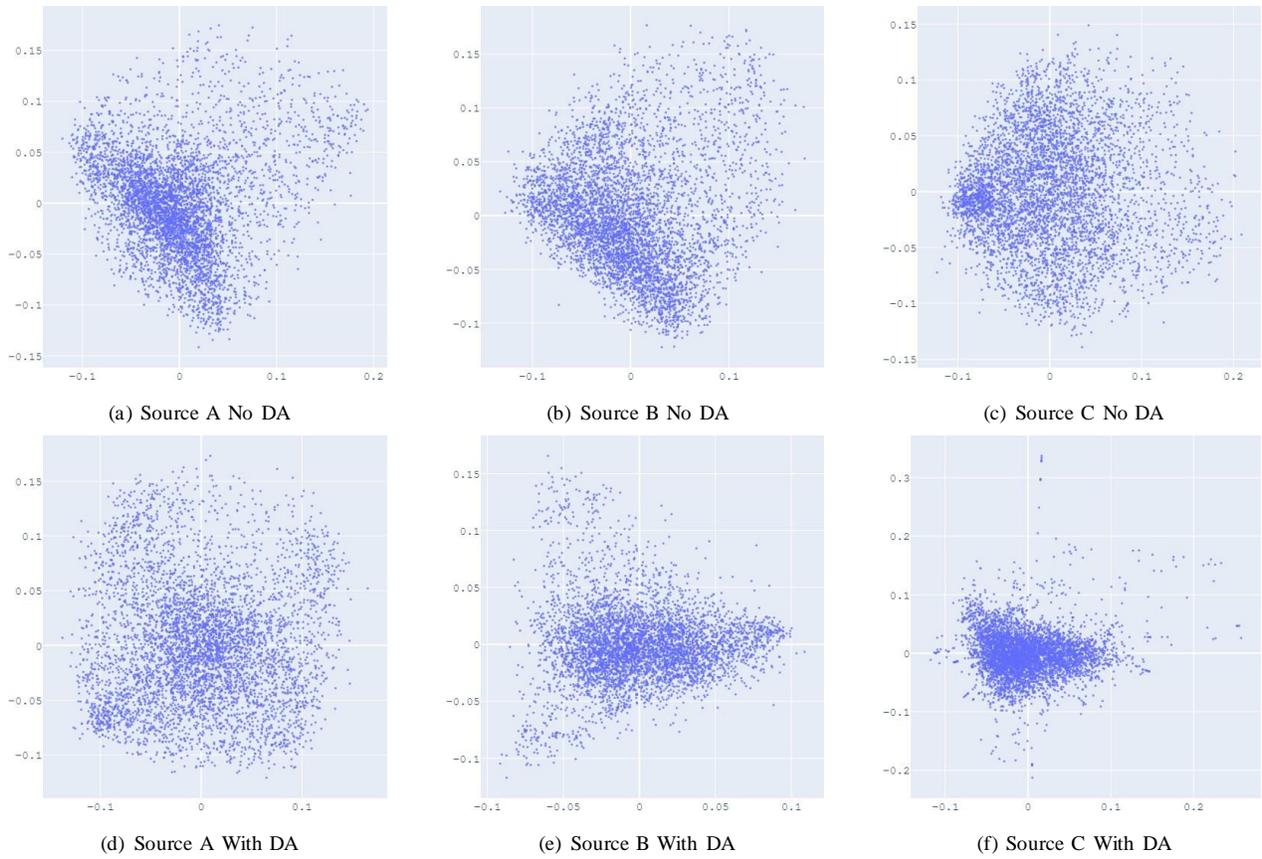

Fig. 5: Isomap plots of the embeddings of WN18 by sources A, B, and C, with and without distribution alignment (DA) respectively. The alignment method used is adversarial regularization.